\documentclass[conference]{IEEEtran}
\usepackage[english]{babel}
\usepackage{amsmath}
\usepackage{graphicx}
\usepackage[colorlinks=true, allcolors=blue]{hyperref}
\usepackage[english]{babel}
\usepackage[utf8x]{inputenc}
\usepackage[ruled,vlined]{algorithm2e}
\usepackage{amsmath}
\usepackage{booktabs}
\usepackage{hyperref}
\usepackage{graphicx}
\usepackage{amsmath,amssymb}
\usepackage{latexsym}
\usepackage{eurosym}
\usepackage{units}
\usepackage{mathtools}
\usepackage{lscape} 
\usepackage{algpseudocode}
\usepackage{amssymb}

\usepackage{microtype}

\title{Reinforcement Learning for Joint V2I Network Selection and Autonomous Driving Policies}
\author{Zijiang Yan, and Hina~Tabassum, {\em Senior Member IEEE}
\\
      Dept. of Electrical Engineering and Computer Science, York University, Canada
}

\begin{document}
\maketitle

\begin{abstract}

Vehicle-to-Infrastructure (V2I) communication is becoming critical for the enhanced reliability of autonomous vehicles (AVs). However, the uncertainties in the road-traffic and AVs' wireless connections can severely impair timely decision-making. It is thus critical to simultaneously optimize the AVs' network selection  and  driving policies in order to minimize road collisions while maximizing the communication data rates. In this paper, we  develop a reinforcement learning (RL) framework to characterize efficient network selection  and autonomous driving policies in a multi-band vehicular network (VNet) operating on conventional sub-6GHz spectrum and Terahertz (THz) frequencies. 
The proposed framework is designed to \textbf{(i)} maximize the traffic flow and minimize collisions by controlling the vehicle's motion dynamics (i.e., speed and acceleration) from autonomous driving perspective, and \textbf{(ii)} maximize the data rates and minimize handoffs  by jointly controlling the vehicle's motion dynamics and network selection from telecommunication perspective. 
We cast this problem as  a Markov Decision Process (MDP) and develop a deep Q-learning based solution to optimize the actions such as acceleration, deceleration, lane-changes, and AV-base station  assignments for a given AV's state. The AV's state is defined based on the velocities and communication channel states of AVs. Numerical results demonstrate interesting insights related to the inter-dependency of vehicle's motion dynamics, handoffs, and the communication data rate. The
proposed policies enable AVs to  adopt
safe driving behaviors with improved connectivity.


\end{abstract}

\section{Introduction}

To support  faster and reliable vehicle to infrastructure (V2I) connections \cite{ko2021v2x}, the emerging  wireless networks will leverage communications on millimeter-waves (mm-waves) and Terahertz (THz) frequencies in conjunction with the conventional sub-6GHz spectrum \cite{rasti2022evolution}. While EHF transmissions can provide  data rates, in the order of multi-Giga-bits-per-second (Gbps),  EHF transmissions are susceptible to severe path-loss attenuation and molecular absorption. Subsequently, frequent switching (or handoffs) among base stations (BSs) due to AV's mobility is expected which can hinder meeting real-time AVs’ communication requirements. Subsequently, incorporating the impact of handoff-related cost  is imperative on the AVs' achievable data rates.


Most of the existing research  on V2I-assisted autonomous driving is focused on collision-avoidance, safe driving, and efficient fuel consumption. In \cite{wu2020driving}, the authors applied a reinforcement learning (RL) framework to arrive  at  the destination  rapidly and the  reward  was proportional to  the vehicle's velocity along with a penalty for vehicle collision.  
In \cite{liu2020reinforcement}, Liu et al. applied double deep Q-network (DDQN)  to enhance the driving safety and fuel consumption of AVs. 
In \cite{szHoke2020driving}, Sz{\H{o}}ke et al. applied deep Q-learning    to control a vehicle in a variety of 
environments. 
The reward was based on the absolute difference between the current and desired velocity.

Other research works focused on enhancing the data rate through radio resource management, regardless of the impact of AV's motion dynamics  on the handoffs and communication data rate. In \cite{ye2019deep}, Ye et al. applied deep reinforcement learning (DRL)  to optimize the power  and sub-band selection. 
The rewards included the data rate and transmission latency of wireless links. In \cite{zhang2019deep}, Zhang et al. applied federated DRL to maximize the data rate and minimize the transmission latency of AVs. 
 In \cite{li2021deep}, Li et al. proposed DDQN algorithm to minimize the energy consumption and computation latency while optimizing the  offloading decisions of AVs. In \cite{hossan2021mobility}, handoff-aware data rate has been characterized in a multi-band RF/THz network using stochastic geometry tools.

To our best knowledge, the existing research have overlooked the interdependency of the AV motion dynamics (velocity, acceleration), handoffs, and wireless data rates. 
To this end, our contributions can be listed as follows:
\begin{itemize}
    \item  We develop a RL-based framework to jointly optimize network selection and autonomous driving policies in multi-band VNets to \textbf{(i)} maximize the traffic flow and minimize collisions, and \textbf{(ii)} maximize the data rates and minimize handoffs   by controlling the AV's motion dynamics and network selection.
    \item The interaction of AV and environment is modeled as a Markov decision process (MDP) characterized by a two dimensional (2D) discrete state-action space for driving actions and network selection actions. Also, we consider a novel reward function that maximizes  data rate and  traffic flow, ensures traffic load balancing across the network, and penalizes handoffs and unsafe driving behaviours.
    \item   Q-Learning and deep Q-network (DQN)-based algorithms are developed. To improve the training efficiency of $Q-$learning, we adopt epsilon decay method  \cite{li_2017}. In the deep $Q-$learning, we consider the Sigmoid as  activation function and adopt mini-batch training from \cite{elfwing2018sigmoid} and\cite{liu2020reinforcement} to improve performance compared with $Q-$learning.
\end{itemize}
The proposed solution enables robust connectivity with improved traffic flow and safety at all times.  
RL  does not require prior network information or access to complete knowledge of the system. Access to such
information is inefficient and even inapplicable for the time varying and uncertain environments. Therefore, the RL-based approach is a promising tool to solve the aforementioned dynamic RRM problem.



\section{System Model and Assumptions}

We consider a two-tier downlink network consist of  $N_R$ RF BSs (RBSs) and  $N_T$ THz BSs (TBSs). We consider a multi-vehicle network with a two-lane road, where $M$ AVs receive information from the BSs (deployed alongside the road) through V2I communications (as depicted in Figure 1). Each AV can associate to only one BS at a time (whether RBS or TBS). The on-board units (OBUs) on the AVs receive real-time information of VNet, i.e., velocity, acceleration, and lane position of the surrounding vehicles.

The  signal transmitted by the RBS is subject to path-loss and short-term  channel fading modeled with Rayleigh distribution. Subsequently, the signal-to-interference-plus noise ratio (SINR) of a $j$-th AV  from BS $i$ can be modeled as \cite{sayehvand2020interference, kaushik2021deep}:
\begin{equation}{\label{sinr_rf}}
    \mathrm{SINR}^{\mathrm{RF}}_{ij} = \frac{P_{R}^{\mathrm{tx}}\:G_{R}^{\mathrm{tx}}\:G_{R}^{\mathrm{rx}} \left(\frac{c}{4\pi f_{R}} \right)^2 H_i }{r_{ij}^{\alpha}\left(N_{R} + I_{R_j}\right)}
\end{equation}
where  $P_{R}^{\mathrm{tx}}, G_{R}^{\mathrm{tx}}, G_{R}^{\mathrm{rx}}, c, f_{R}, r_{ij},$ and $\alpha$ denote the transmit power of the RBSs, antenna transmitting gain, antenna receiving gain, speed of light, RF carrier frequency (in GHz), distance between the $j$-th AV  and the $i$-th RBS, and path-loss exponent, respectively. 
Also, $H_i$ is the exponentially distributed channel fading power observed at the AV from the $i$-th RBS,  $N_R$ is the power of thermal noise  at the receiver, $I_{R_j} = \sum_{k\neq i} P_{R}^{\mathrm{tx}} \gamma_{R} r_{kj}^{-\alpha} H_{k}$  is the cumulative interference at $j$-th AV from the interfering RBSs. From the cumulative interference, $r_{kj}$, is the distance between the $k$-th interfering RBS and the $j$th AV, $H_{k}$ is the power of fading from the $k$-th interfering RBS to the typical AV, and $ \gamma_R = G_{R}^{\mathrm{tx}} \: G_{R}^{\mathrm{rx}} \left( {c}/{4\pi f_{R}} \right)^2 $. 

In THz network, due to the presence of molecular absorption, the  line-of-sight (LOS) transmissions are far more significant than the non-line-of-sight (NLOS) transmissions. Therefore, the SINR of a $j$th AV can be modeled as:
\begin{align*}{\label{sinr_1}}
    \mathrm{SINR}^{\mathrm{THz}}_{ij} &= \frac{G_{T}^{\mathrm{tx}} G_{T}^{\mathrm{rx}} \left(\frac{c}{4\pi f_{T}} \right)^2 P_{T}^{\mathrm{tx}}\: \mathrm{exp}(-K_a(f_T) r_{ij})r_{ij}^{-2} }{N_{T_j} + I_{T_j}},
   \end{align*}
where $ G_{T}^{\mathrm{tx}}, G_{T}^{\mathrm{rx}},P_{T}^{\mathrm{tx}}, f_{T}, r_{ij},$ and $K_{a}(f_T)$ denote antenna transmitting gain of the TBS,  antenna receiving gain of the TBS, the transmit power of the TBSs, THz carrier frequency, distance between the $j$th-AV to the  $i$th-TBS, and the molecular absorption coefficient relies on the composition of the medium and also on the frequency (i.e., $f_{T}$) of the signal, respectively\footnote{For the sake of brevity, we will drop the argument of $K_a(f_T)$ from this point onwards in the paper.}. Note that $G_{T}^{\mathrm{rx}}\left(\theta\right)$ and $G_{T}^{\mathrm{tx}}\left(\theta\right)$ are receiver antenna and directional transmitter gains, respectively. The  beamforming gains from the main lobe and side lobes of the TBS transmitting antenna can be defined in below:   
\begin{equation}
\label{eq:gain2}
  G_{T}^{\mathrm{q}}\left(\theta\right) =
    \begin{cases}
      G^q_{\mathrm{max}} & \mid \theta \mid \leq w_{q}\\ 
      G^q_{\mathrm{min}} & \mid \theta \mid > w_{q}
    \end{cases},  
\end{equation}
where $q\in \{\mathrm{tx,rx}\}$, $\theta \in [-\pi,\pi)$ is the boresight direction angle, $w_{q}$ is the beamwidth of the main lobe, $G^q_{\mathrm{max}}$ and $G^q_{\mathrm{min}}$ are the beamforming gains of the main and side lobes, respectively. Assuming the typical AV's receiving beam aligns with the transmitting beam of the associated TBS through beam alignment techniques. However, for the alignment between the user and interfering TBSs, we define a random variable $D$, which can be generalized as  $D \in \{G^{\mathrm{tx}}_{\mathrm{max}}G^{\mathrm{rx}}_{\mathrm{max}},G^{\mathrm{tx}}_{\mathrm{max}}G^{\mathrm{rx}}_{\mathrm{min}},G^{\mathrm{tx}}_{\mathrm{min}}G^{\mathrm{rx}}_{\mathrm{max}},G^{\mathrm{tx}}_{\mathrm{min}}G^{\mathrm{rx}}_{\mathrm{min}}\},$ and the respective probability for each case is $F_{\mathrm{tx}}F_{\mathrm{rx}}$, $F_{\mathrm{tx}}(1-F_{\mathrm{rx}})$, $(1-F_{\mathrm{tx}})F_{\mathrm{rx}}$, and $(1-F_{\mathrm{tx}})(1-F_{\mathrm{rx}})$, where $F_{\mathrm{tx}} = \frac{\theta_{\mathrm{tx}}}{2\pi}$ and $F_{\mathrm{rx}} = \frac{\theta_{\mathrm{rx}}}{2\pi}$. 
     Without loss of generality, we consider negligible side lobe gains.
     Subsequently, $I_{T} = \sum_{k\neq i}\gamma_T \:P_{T}^{\mathrm{tx}}\:F_{\mathrm{tx}}F_{\mathrm{rx}} r_{kj}^{-2} \mathrm{exp}(-K_a \:{r_{kj}})$ is the cumulative interference at the AV  and $r_{kj}$ is the distance of $j$th-AV to the $k$-th interfering TBS. The cumulative thermal and molecular absorption noise is thus given as:
$
    N_{T_j}= N_{0} +  \:P_{T}^{\mathrm{tx}} \gamma_T \:{r_{ij}^{-2}} \:(1-e^{-K_a \:{r_{ij}}})+
    \sum_{k \neq i} \gamma_T F_{\mathrm{tx}}F_{\mathrm{rx}} \:P_{T}^{\mathrm{tx}} \:{r_{kj}^{-2}}(1-e^{-K_a \:{r_{kj}}}),
$
where $\gamma_T=G_{T}^{\mathrm{tx}} G_{T}^{\mathrm{rx}} \left(\frac{c}{4\pi f_{T}} \right)^2$.
\begin{figure}
\includegraphics[width=1\linewidth]{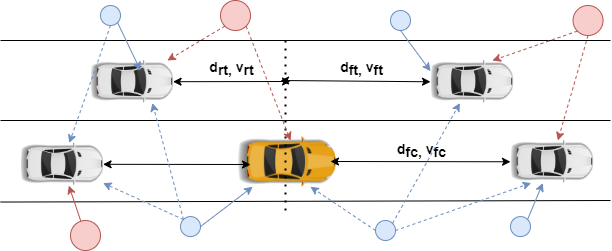}
\caption{An illustrative structure of  the multi-band VNet model. The blue and red circles represent TBSs and RBSs, respectively. The solid and dash line represent desired signal links and interference links, respectively.}
\label{fig:BS_distributions}
\end{figure}

 Each RBS and TBS has the available bandwidth given by $W_R$ and $W_T$, respectively.  All AVs are equipped with a single antenna. Each AV measures the channel quality as well as signal and interference levels from each BS.  Subsequently, the data rate of each AV to BS link  can be computed as $T_{ij} = W_j\log_2(1+\mathrm{SINR}_{ij})$, where $W_j$ is the transmission bandwidth of the BS $j$. Each RBS and TBS has a maximum limit of $Q_R$ and  $Q_T$ on the number of AVs that can be supported, respectively.  Each AV maintains a list of top three BSs in terms of the achievable data rate and then informs those BSs.  Consequently, each BS can calculate the number of possible AV associations at each time instance denoted by $n_c$. As AVs drive along the corridor, they switch from (connecting to) one BS to another, which is called a ``handoff\footnote{We consider two types of handoffs: horizontal and vertical. Horizontal handoff refers to the situation where the AV connection moves from one BS to another BS of the same type. Vertical handoff refers to the situation where the AV connection moves from one specific type of BS to a different type of BS, such as moving from RBS to a TBS.}". 
Frequent handoffs can severely impact the received AV data rate due to handoff latency/ failures.   We discourage handoffs by introducing a handoff penalty ($\mu$) which is higher for TBS and lower for RBS because THz transmission  is prone to short communication distances.

\section{MDP Formulation and Reinforcement Learning Algorithms}
\label{section:mdpandalgorithm}
In this section, we specify the states of each AV, 2D action space, and rewards of the considered problem. Then, we present the RL algorithm.

\subsection{Observation and State Space}

First, we introduce an observation space to model the surrounding environment.
The observation of a given AV includes eight variables, i.e., \textbf{(1)}  relative distance between the given AV and  the front AV in the current lane in $x$ direction, $d_\mathrm{fc}$, \textbf{(2)}   relative distance between the given AV and  the front AV in the adjacent lane in $x$ direction, $d_\mathrm{ft}$, \textbf{(3) } relative distance between the given AV and  the rear AV in the adjacent lane in $x$ direction, $d_\mathrm{rt}$, \textbf{(4)} relative velocity between the given AV and  the front AV in the current lane in $x$ direction, $v_\mathrm{fc}$, \textbf{(5) } relative velocity between the given AV and  the front AV in the adjacent lane in $x$ direction, $v_\mathrm{rc}$, \textbf{(6)}  relative velocity between the given AV and  the rear AV in the adjacent lane in $x$ direction, $v_\mathrm{rt}$, \textbf{(7)}~number of BSs that satisfy the desired data rate of the given AV among the three  BSs offering best data rates around the current AV, $c$, \textbf{(8)}  lane-id in which a given AV stays, $l_\mathrm{id}$.


We discretize the distance observation variables $d_{\mathrm{fc}},d_{\mathrm{ft}},d_{\mathrm{rt}}$ into three states (i.e., $d \leq d_c, d_c \leq d \leq d_f, d>d_f$), where $d \in \{d_{\mathrm{fc}},d_{\mathrm{ft}},d_{\mathrm{rt}}\}$, $d_c$ is the minimum safety distance, and $d_f$ is the maximum safety distance. Similarly, we discretize velocity observation variables $v_{\mathrm{fc}},v_{\mathrm{ft}},v_{\mathrm{rt}}$ into three states (i.e., $v < 0$ if two AVs are approaching each other, $v=0$ if two AVs have the same velocity, $v>0$ if two AVs are moving apart), where $v$ is the relative velocity between the two AVs, and $v \in \{v_{\mathrm{fc}},v_{\mathrm{ft}},v_{\mathrm{rt}}\}$. Now the number of BSs denoted by $c$ out of three nearest BSs who can support the data rate requirements  of a given AV are represented by three states (i.e., $0\leq c\leq~1, c=2, c=3$). Finally, the lane observations are discretized into two states for right and left lane driving. The state can be expressed as 
\begin{equation*}
    \mathrm{s}_t = \{d_{\mathrm{fc}}(t), d_{\mathrm{ft}}(t), d_{\mathrm{rt}}(t), v_{\mathrm{fc}}(t), v_{\mathrm{ft}}(t), v_{\mathrm{rt}}(t), c(t), l_{\mathrm{id}}(t)\}
\end{equation*}

\subsection{Two Dimensional Action space}

At each time step $t$, each AV selects a driving-related action $a_t^\mathrm{tran}$ and telecommunication-related action $a_t^\mathrm{tele}$ from the action space $\mathcal{A}$. The action  space consists of self-driving action space $\mathcal{A}_\mathrm{tran}$ and telecommunication action space $\mathcal{A}_\mathrm{tele}$, i.e., $\mathcal{A} = \{  \mathcal{A}_\mathrm{tran},  \mathcal{A}_\mathrm{tele}\}$. For each time step, the AV must select both  $a_t^\mathrm{tran}$ and   $a_t^\mathrm{tele}$. We have seven discrete actions from the self-driving side and, for each self-driving related action, we have three telecommunication related actions. Thus, the total number of actions are 21. The actions include:

\subsubsection{Self-Driving Actions} are listed as follows.
\textbf{(1)}~\textit{Hard Acceleration:} when the AV increases speed $v +v_{at}$, where $v_{at}$ is the additional target speed, $a_x =4$~m/s$^2$,
\textbf{(2)}~\textit{Mild Acceleration:} when the AV increases speed $v +v_{at}$,   $a_x =1.5$~m/s$^2$,
\textbf{(3)} \textit{Maintains:} the AV maintains the  speed $v$, $a_x=0$,
\textbf{(4)} \textit{Mild Deceleration:} the AV decreases speed $v -v_{at}$, i.e., $a_x =-1.5$~m/s$^2$,
\textbf{(5)}  \textit{Hard Deceleration:} the AV decreases speed $v -v_{at}$, i.e., $a_x =-4$~m/s$^2$,
\textbf{(6)} \textit{Lane Switch:}  We define $v_y$ as the instant vertical velocity of AV if it switches the lane. AV switches to the adjacent lane if $v_y = 1.5$~m/s, otherwise, $v_y = 0$~m/s and AV remains in the lane where it is currently in,
\textbf{(7)} \textit{Stop:}  AV will slow down to 0~m/s instantly (i.e., $v_x, v_y= 0$m/s) and $a_x= -0.85 v_x/dt$, where $v_x$ is the velocity of AV in the previous time step and  $dt$ is the duration of time step  when AV collides.

\subsubsection{Telecommunication Actions} are listed as follows.
\textbf{(1)} \textit{Handoff-aware Network Selection:} Each AV prepares a sorted list of three BSs offering best data rates and associates to those who can fulfill data rate requirement of the AV $R_{\mathrm{th}}$. Then, the AV collects the traffic load information from these three BSs (i.e.,  the number of AVs associated with each BS $n_s$). Based on the quota of each BS $j$, $Q_j \in [Q_R, Q_T]$, each AV computes a \textit{weighted data rate metric} that encourages traffic load balancing at each BS and discourages unnecessary handoffs, i.e.,
    $$ T_q = \frac{T_{ij}}{\min \left(Q_j, n_s \right)} (1 - \mu)$$
   where $\mu$ denotes the handoff penalty to discourage unnecessary handoffs, especially in TBSs.
                    \begin{equation}
                            {
                            \mu = 
                        \begin{dcases}
                            0.5 ,& \text{if switch to a TBS} \\
                            0.1, & \text{if switch to a RBS } \\
                            0, & \text{Keep Previous BS} 
                        \end{dcases}
                            }   
                    \end{equation}
Each AV then  connects to the BS with  maximum ${T}_{{q}}$.
\textbf{(2)}~\textit{Network Selection with No Handoff Consideration:}
The AV computes $T_q$ by substituting $\mu=0$ and prepares a sorted list of three top BSs and choose to connect to BS with the maximum  ${T}_{{q}}$.
\textbf{(3)} \textit{Network Selection based on Maximum Data Rate:}
The AV computes $T_q = T_{ij} $ and chooses to connect to a BS with the maximum data rate.

\subsection{Rewards}
The design of the associated reward/penalty function is directly related to optimizing the driving policy and network selection. The reward function is critical for accelerating the
convergence of the model. When the
AV is receiving a higher handoff-aware data rate, while guaranteeing
safe driving, it receives a positive reward. By taking any other
actions, which may lead to an increase of the handoffs, collision or traffic violation, the AV receives a penalty. 

\subsubsection{Autonomous Driving Rewards}
We define the reward function from the driving side as follows:
$$
    r^{\mathrm{tran}}_{t} = w_1 c + w_2 v +w_3 h +w_4 a +w_5 l
$$
where $c \in [-1,0]$ denotes the collision indicator, $v = 0.2 \,(v^t_x -v_{\mathrm{desired}})$ is the normalized AV velocity $v \in [-1,1] $ with a scaling factor 0.2 and  the expected traffic velocity  $v_{\mathrm{desired}}$, and $h \in [-1,0,1]$ indicates whether the AV of interest is close, medium, and far from the front AV which  represents the traffic congestion, normal traffic, and free-way traffic, respectively.
Also, $a \in [-3,-1,0]$  is the penalty for the driving acceleration behaviours. Specifically, if the AV chose to accelerate or decelerate hard, then $a = -3$. On the other hand, if the AV choose to accelerate or decelerate mild, then $a = -1$. Otherwise, we penalize it as $a = 0$. Finally, $l \in [-1,0]$ is the penalty factor for lane. We penalize AV on the left lane and encourage AV on the right lane to avoid speeding.

\subsubsection{Telecommunication Rewards}
We define the reward from telecommunication side as follows:
\begin{equation}
    r^{\mathrm{tele}}_{t}= w_6 T_q (1- \text{min}(1,k))
\end{equation}
where  $T_q$  is the handoff aware data rate and  $k$ is the handoff probability computed by dividing the number of handoffs with the current time steps $t$. 

\textit{Remark:} Note that $w_1 \dots w_6$ are the  weights to set the priority of each term. For instance, $w_1$ needs to be sufficiently large compared to $w_2, \cdots, w_6$ for collision avoidance. Also, $w_6 \ll w_2, w_3, \cdots, w_5$ to normalize the high values of communication data rates.

To obtain  a reasonable performance, we consider intermediate reward  and find the policy $\mathcal{J}$ to maximize the discounted expected cumulative rewards.
\begin{equation}
    R_t = \mathbb{E}[\sum_{t=0}^{\infty}\gamma(r^{\mathrm{tran}}_t + r^{\mathrm{tele}}_{t})]
\end{equation}
where $\gamma \in [0,1]$ is the discount factor.
The state transitions and rewards are a function of the  AV  environment and actions taken by the AV. At time step $t$, the transition from state $s_t$ to $s_{t+1}$ can be defined as the conditional transition probability given by $p(s_{t+1},r_t|s_t,a_t^\mathrm{tele},a_t^\mathrm{tran})$. Note that the AVs do not have any prior information of the transition probabilities. 


\subsection{Q-Learning  Algorithm}
Each AV  interacts with the environment to observe the system state $s_t$ at every time-step $t$ from the state space $\mathcal{S}$. 
Then, each AV chooses an appropriate action $a_t$ from the two-dimensional action space $\mathcal{A}$ selecting the optimal driving and network selection choice based on policy, $\mathcal{J}$. 
\begin{equation}
    s_t \in  \mathcal{S} \rightarrow a^{\mathrm{tele} }_{t} \in \mathcal{A}_\mathrm{tele},  a^{\mathrm{tran} }_{t} \in \mathcal{A}_\mathrm{tran}
\end{equation}
The decision policy
$\mathcal{J}$ is determined by a Q-table, $Q(s_t, a_t)$, where Q-value  represents the performance of a specific action in the given state. The  policy at each time slot is to choose a specific action that results in the maximum Q-value, i.e.,
\begin{equation}
    a_t = \arg \max_{a \in \mathcal{A}} Q(s_t,a)
\end{equation}
Given the chosen action,
the AV state traverses to a new state $s_{t+1}$ yielding a reward, $r_t$. We apply the $\epsilon-$greedy search strategy to enable the convergence of Q-table to an optimal Q function, where a fixed $\epsilon$ provides the constant exploration rate in \cite{kochenderfer2015decision}. To boost the exploration efficiency in the initial stage, $\epsilon$ annealing method was applied where $\epsilon$ decays  gradually with the iterations (from 1 to 0).  To obtain the optimal policy with optimal Q values, we  use Bellman update, i.e.,
\begin{equation*}
Q(s_{t},a_t) \gets Q(s_{t},a_t) + 
  \alpha(r_t + \gamma \mathrm{max}_{\alpha}Q(s_{t+1},a_t)-Q(s_{t},a_t)).
\end{equation*}
The  $Q-$value is generated for each state-action pair at each time step. We incorporate the learning rate $\alpha$ to handle the bias between the current Q-value $Q(s_{t},a_t)$ and the future Q-value $Q(s_{t+1},a_t)$.
The higher value of $\alpha$ puts more emphasis on $Q(s_{t+1},a_t)$. The algorithm is presented in
\textbf{Algorithm~1}.
\begin{algorithm}
\SetAlgoLined
\KwResult{Optimal state-action function $Q^*(s,a)$ }
 Initialization $Q(s,a) \gets 0$
 \\
 \While{$\mathrm{episode} < \mathrm{episode \ limit \ or \ run \ time}< \mathrm{time \ limit}$ }{
  $t \gets 0$\;
  $s_t \gets$ Random Initialization \;
  \While{$t \leq \mathrm{horizon \ limit} $ }{
  \eIf{$rand() \leq \epsilon $}{
   Randomly choose $a^{\mathrm{tran} }_{t}$ from action space $a^{\mathrm{tran} }_{t}$\;
   Randomly choose $a^{\mathrm{tele} }_{t}$ from action space $a^{\mathrm{tele} }_{t}$\;
   }{
   $a_t = \mathrm{argmax}_{\alpha} Q(s_t,a)$\;
   $a^{\mathrm{tran} }_{t} \gets a_t(1)$\;
   $a^{\mathrm{tele} }_{t} \gets a_t(2)$\;
  }
 Apply $a^{\mathrm{tran} }_{t}$ and $a^{\mathrm{tele} }_{t}$ to AV\;
  $r_t \gets $ reward function$(s_t)$\;
  $s_{t+1} \gets $ model based on $s_t$ and $a_t$\;
   $Q(s_{t},a_t) \gets  Q(s_{t},a_t) +\alpha(r_t + \gamma \mathrm{max}_{\alpha} Q(s_{t+1},a_t)-Q(s_{t},a_t))$\;
  $t \gets t + 1 $
  }
 $Q^*(s,a) \gets Q^(s,a) $ 
 }
  \caption{Q-learning Algorithm to Optimize Autonomous Driving and Network Selection Policies}
\end{algorithm}

\subsection{Deep Q-Learning Algorithm}
\label{subsection:deepq}
To improve the performance of Q-learning, we consider a fully-connected feed-forward neural network (FNN) $N(s)$ with weights $\{\theta\}$ to approximate the Q-value for a given action and state \cite{ye2019deep}. FNN takes the state as input and output the Q-value for every action.
Since Q-values are real, the FNN performs a multivariate linear regression task. We use  ReLU activation function ($f_r(x) = \mathrm{max}(0,x)$) as the first layer, Sigmoid activation function ($f_r(x) = \frac{1}{1+e^{-x}}$) for hidden layers,  and linear activation function for the output layer. 
We adopt Binary Cross-Entropy Loss function defined as:
\begin{equation}
    L = -(y \log{\hat{y}} + (1-y) \log(1-\hat{y}))
\end{equation}
where $\hat{y}$ is target value and $y$ is the real value.
Due to the correlation
of input states, over-fitting issues and convergence to a local optimum may occur. Therefore, we suggest  experience replay method \cite{chen2022deep} to stabilize the NN's output.
\begin{enumerate}
\item \textit{Target network:}  We create a new Neural Network to compute targets $\hat{y} = \hat{Q}(s',a';\theta)$. For each 50 episodes, we substitute network $\hat{N}$ with the network $N$. The objective of target network is providing the stable $Q$ values during training.
\item \textit{Experience Replay:} Gathering previous information from telecommunication side and transportation side. We require $(s,a^{\mathrm{tele} }_{t},a^{\mathrm{tran} }_{t},s',r)$ to store in temporary memory. Mini-batches methods \cite{hou2017novel} are applied to the past learning experience to reduce correlation. 

\end{enumerate}

\begin{algorithm}
\SetAlgoLined
\KwResult{Action function $Q_\theta$ and Policy $\mathcal{J}$ }
\KwData{$Q$-network, Experience replay memory $D$, mini batch-size $m$}
\textbf{Initialization:} $\mathcal{D} \gets \bold{0}$,  Q-network weights $\theta \gets \bold{0}$, Target network $\theta^* \gets \theta$, $Q(s,a)$, AVs, TBSs, RBSs 
\While{$\mathrm{episode} < \mathrm{episode \ limit \ or \ run \ time}< \mathrm{time \ limit}$}{
  $t \gets 0$,
  $s_t \gets \mathrm{horizon \ limit} $\\
\While{$t \leq \mathrm{horizon \ limit} $}{AV selects $a_t$ by $\epsilon$-greedy search as  \textbf{Algorithm-1}.\\  
Derive $ a^{\mathrm{tran} }_{t}$ and $a^{\mathrm{tele} }_{t}$  from $a_t$\\
Apply $a^{\mathrm{tran} }_{t}$ and $a^{\mathrm{tele} }_{t}$ to AV;\\
Compute reward $r_t$ and update $s'$;\\
Store $(s_t,a_t,s_{t+1},r_t)$ to $\mathcal{D}$;\\
\textbf{Experience Replay:} sample transitions mini-batch in $\mathcal{D}$ $(s_k,a_k,r_k,s'_k)$ where  $k \in m$;\\
\textbf{Set target-$Q$ function:} $ \hat{y}_k = r_k + \gamma\max_{a'}\hat{Q}(s'_k,a'_k;\theta_k) $;\\
\textbf{Set real $Q$-function:} $ y_k = {Q}(s_t,a_t;\theta)$;\\
Compute loss \\$ \mathcal{L}(\theta) = \sum_{k \in m} -y_k \log{\hat{y}_k} + (1-y_k) \log(1-\hat{y}_k)$;\\
Perform gradient descent step by minimizing loss $\mathcal{L}$;
$\theta \gets \theta - a_{t} \cdot \mathcal{L}(\theta) \cdot \triangledown_{\theta}{y_k}$;\\
Update deep Q network weights $\theta \gets \theta^*$
  }
   Policy $\mathcal{J}$ updated in terms of Q
 }
  \caption{Deep Q-learning Algorithm to Optimize Network Selection and Autonomous Driving Policies}
\end{algorithm}

\section{Numerical Results and Discussions}
In this section, we demonstrate the performance of the proposed  algorithms and highlight the complex  dynamics between the wireless  connectivity, traffic flow, and AV's speed. 
 
 
 Unless stated otherwise, we use the following simulation parameters. We consider 4 RBSs operating on 3.5 GHz and 10 TBSs operating on 1 THz.  We assume RBS quota $Q_R =2$ and TBS quota $Q_T = 5$. Note that $G_{T}^{\mathrm{tx}} =G_{T}^{\mathrm{rx}} = 316.2$. The transmission bandwidth of each TBS and RBS is  $W_T = 5 \times 10^8$ and  $W_R =4 \times 10^7$, respectively.
 The weights of the reward functions are selected as $w_1$ = 1000, $w_2=5, w_3=w_4=w_5=1, w_6 = 4.5 \times 10^{-6.5}$.
The transmit powers of TBSs and RBSs are $P_{R}^{\mathrm{tx}}$ = 1~W and $P_{T}^{\mathrm{tx}}$ = 1~W, respectively. Also, $K_a$ = 0.05~m$^{-1}$ and the path-loss exponent $\alpha=4$.

\begin{figure*}
\centering
\begin{tabular}{lccccc}
\includegraphics[width=0.3\linewidth]{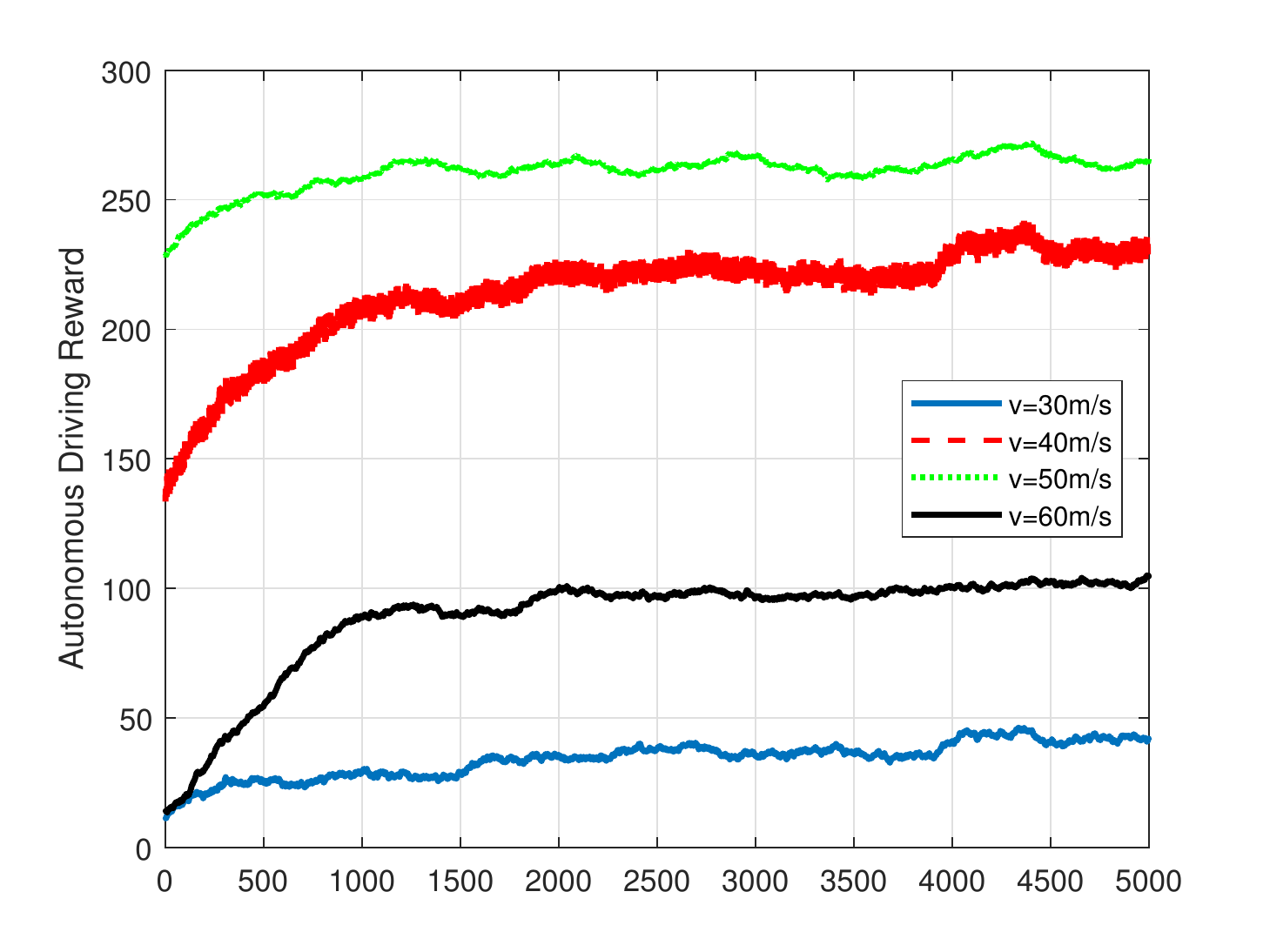}\hspace{-1cm}&\hspace{0.25cm}
\includegraphics[width=0.3\linewidth]{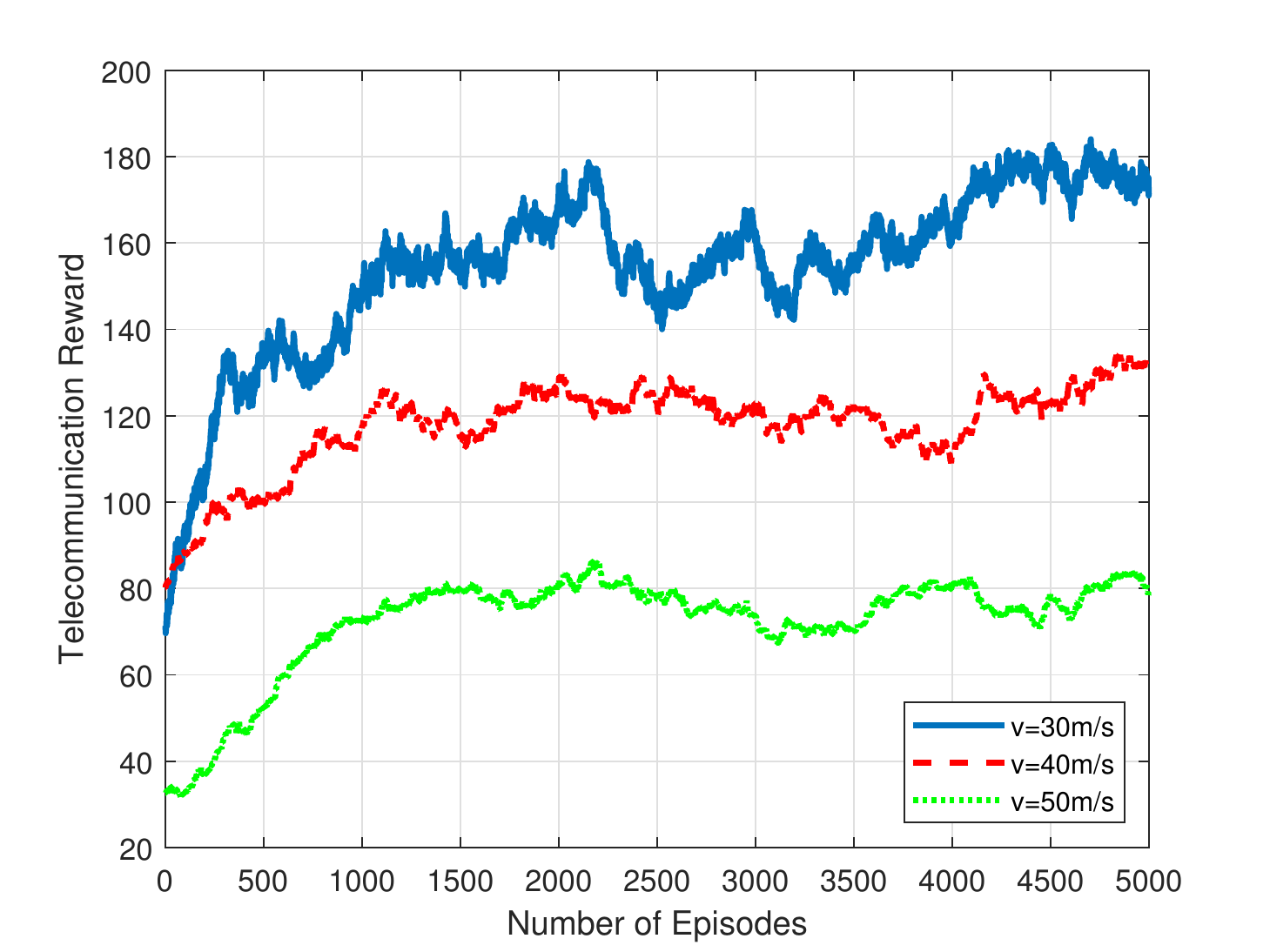}\hspace{-1cm}&\hspace{0.25cm}
\includegraphics[width=0.3\linewidth]{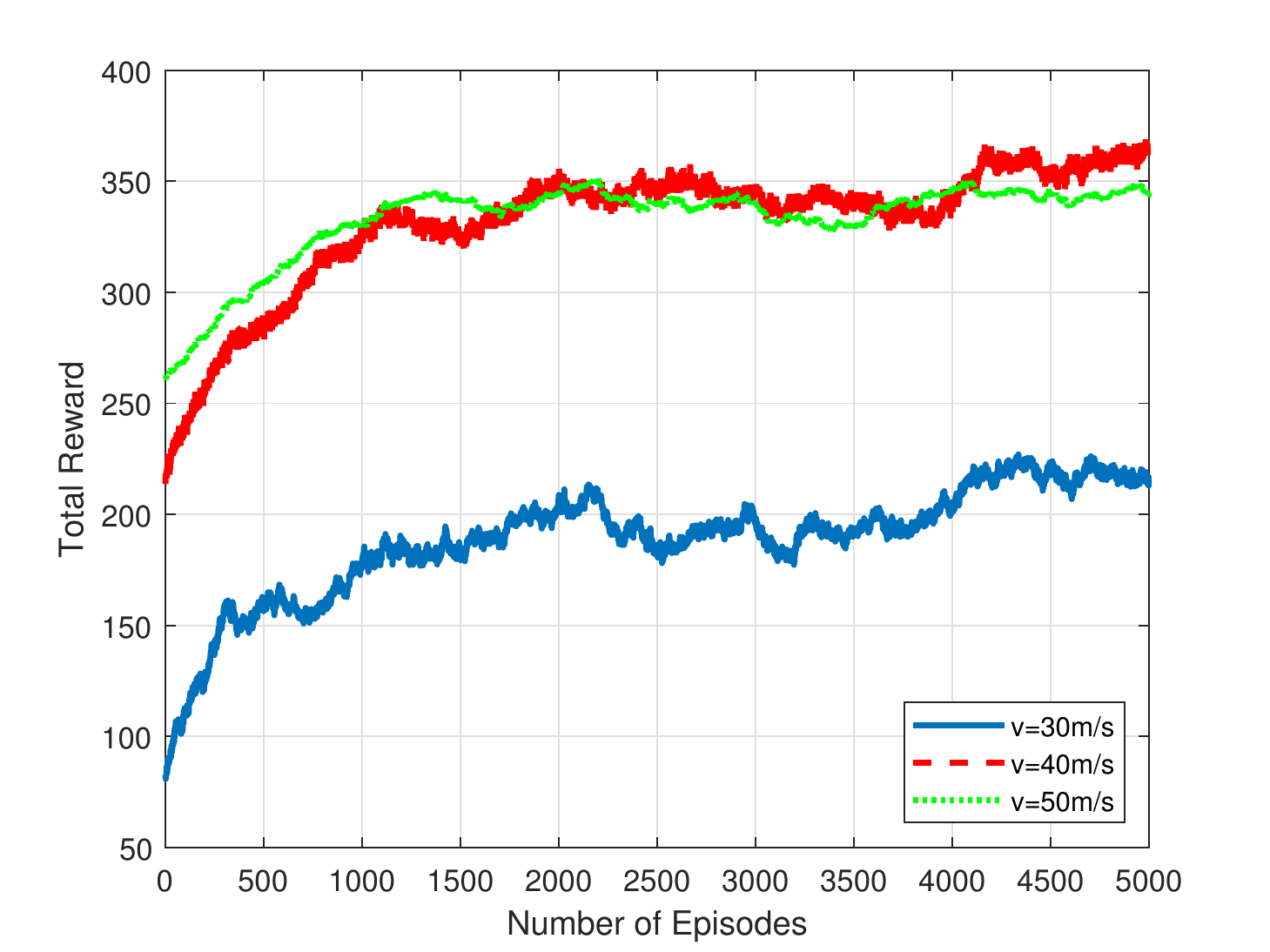}&\\
\qquad \qquad\qquad \qquad (a)  & (b) & (c) 
\end{tabular}
\vspace{-2mm}
\caption {(a) Autonomous driving rewards, (b) Telecommunication rewards, and (c) Total rewards with varying velocities of AV.} 
\label{fig:qplot}
\end{figure*}

Fig.~\ref{fig:qplot} depicts the  self-driving reward, telecommunication reward, and the total reward as a function of the desired velocity of the AV. We note that the transportation reward in Fig.~\ref{fig:qplot}(a) increases with the increase in the desired AV velocity (from 30 m/s to 50 m/s) due to improved traffic flow; whereas, the telecommunication reward in Fig.~\ref{fig:qplot}(b) decreases with the increase in velocity due to handoffs. However, a critical note in Fig.~\ref{fig:qplot}(a) is that increasing speed increases the collision probabilities which again reduces the self-driving rewards. Finally, Fig.~\ref{fig:qplot}(c) suggests that setting a desired velocity of 40~m/s - 50~m/s is favourable as it provides a comparable overall utility.  It can be observed that the average
AV rewards  for each episode increases
with the training episodes which demonstrates the convergence of the proposed algorithm.

\begin{figure*}
\vspace{-1mm}
\centering
\begin{tabular}{lccccc}
\includegraphics[width=0.3\linewidth]{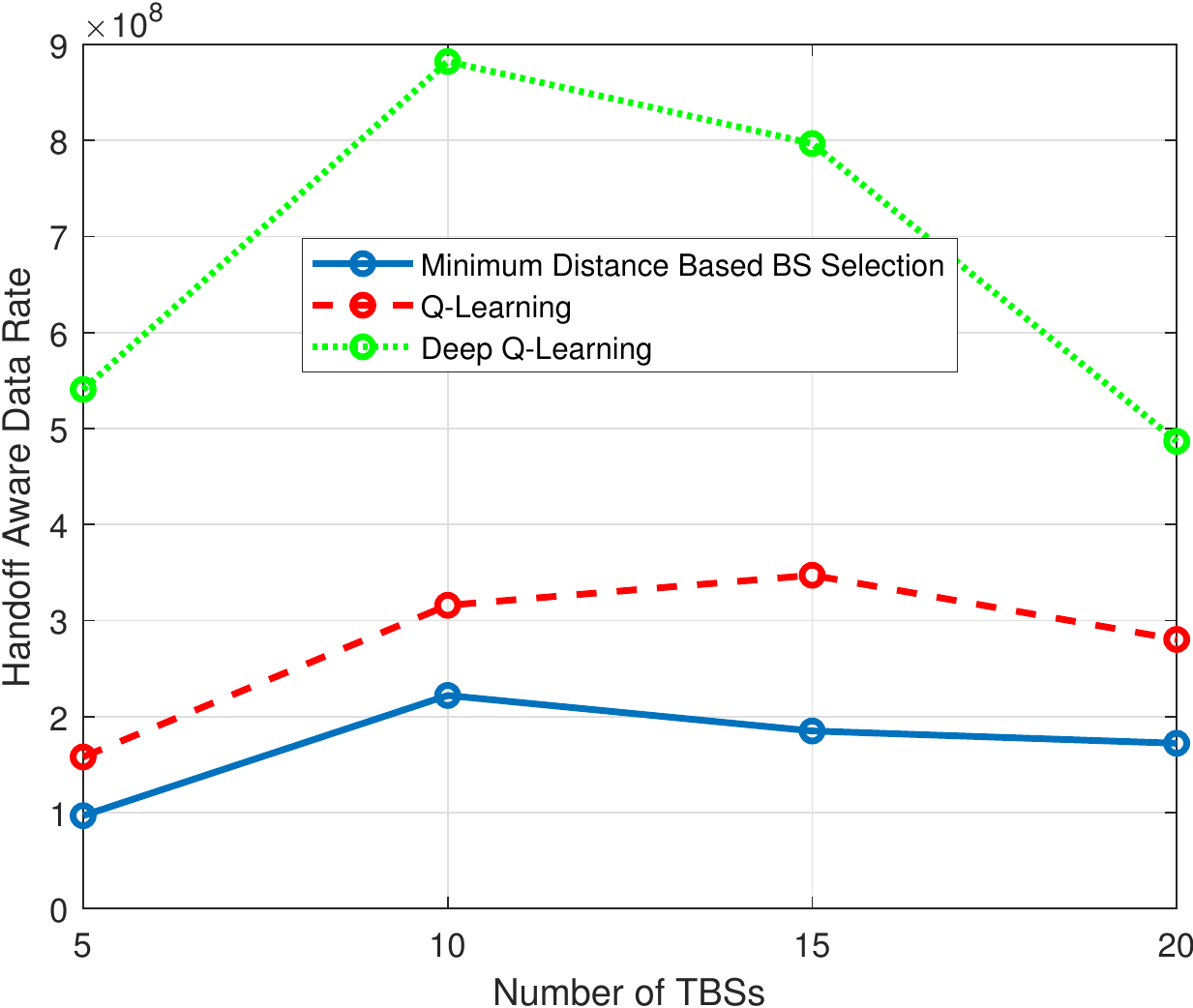}\hspace{-1cm}&\hspace{0.25cm}
\includegraphics[width=0.3\linewidth]{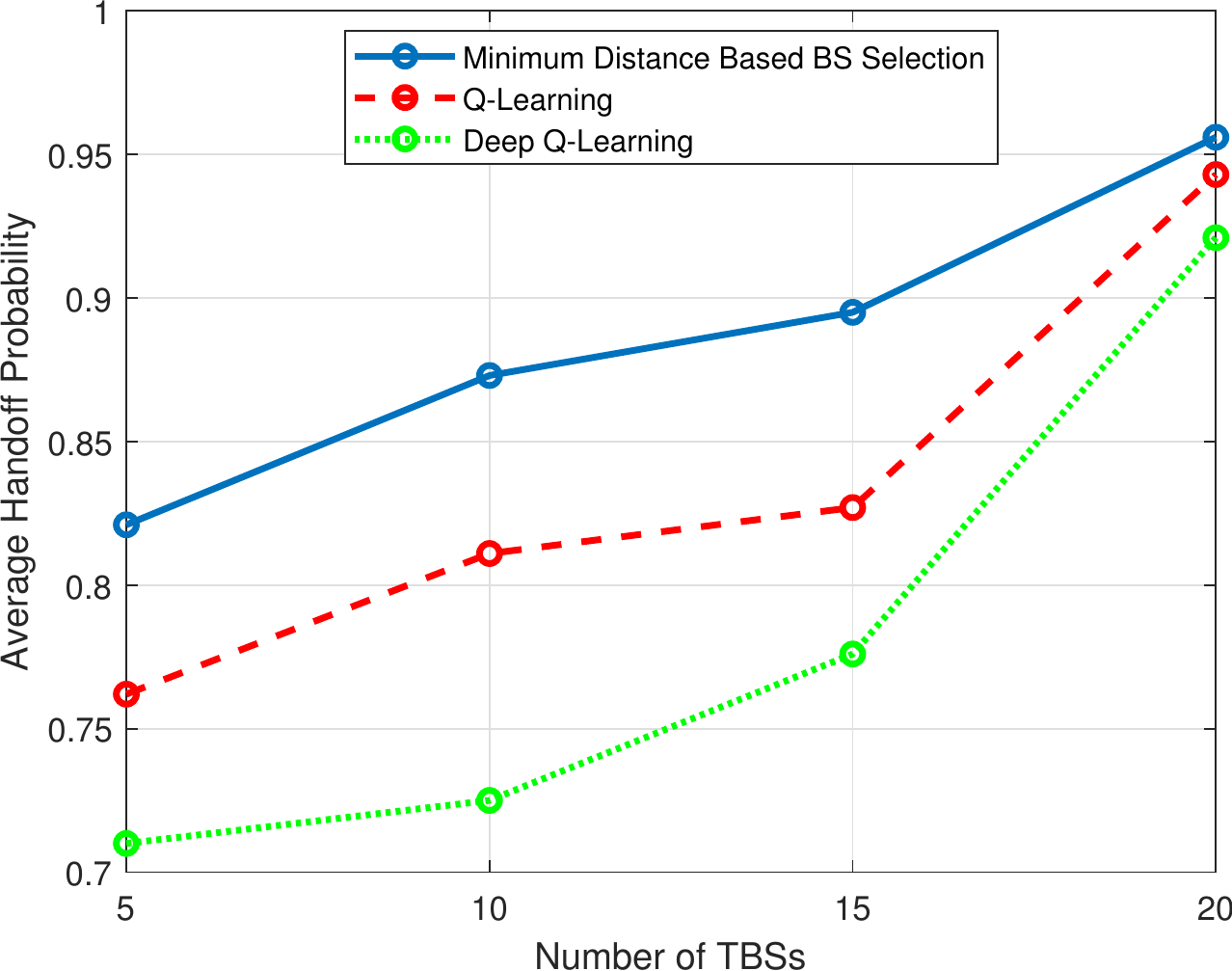}\hspace{-1cm}&\hspace{0.25cm}
\includegraphics[width=0.3\linewidth]{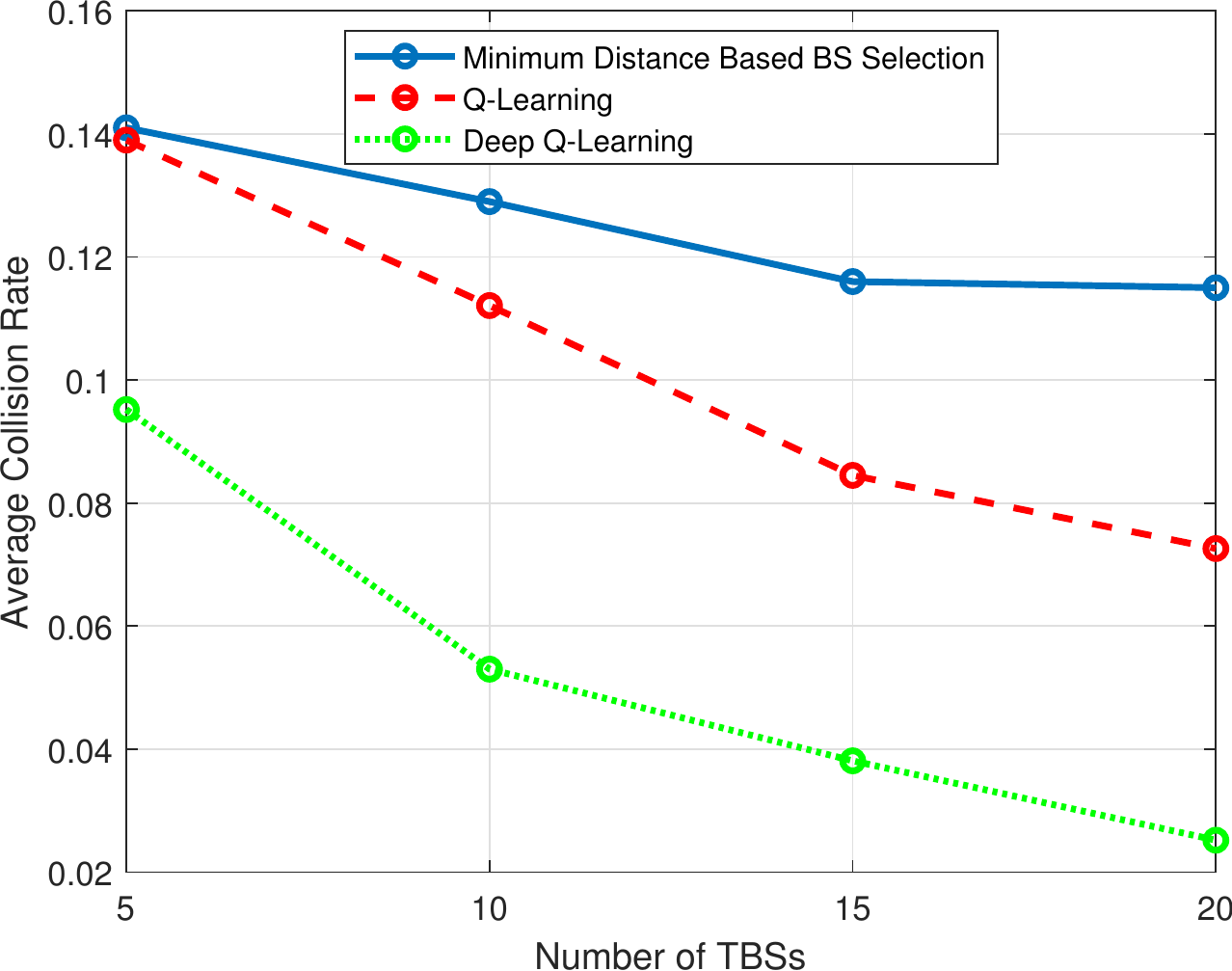}\hspace{-1cm}&\hspace{0.25cm}&\\
\qquad \qquad\qquad \qquad (a)  & (b) & (c) 
\end{tabular}
\vspace{-2mm}
\caption {(a) Handoff-aware data rate (b) Average handoff probability, (c) collision rate  considering   5 RBSs, 25 AVs with their desired velocities $v =30$m/s.} 
\label{fig3}
\end{figure*}

\begin{figure*}
\vspace{-1mm}
\centering
\begin{tabular}{lccccc}
\includegraphics[width=0.3\linewidth]{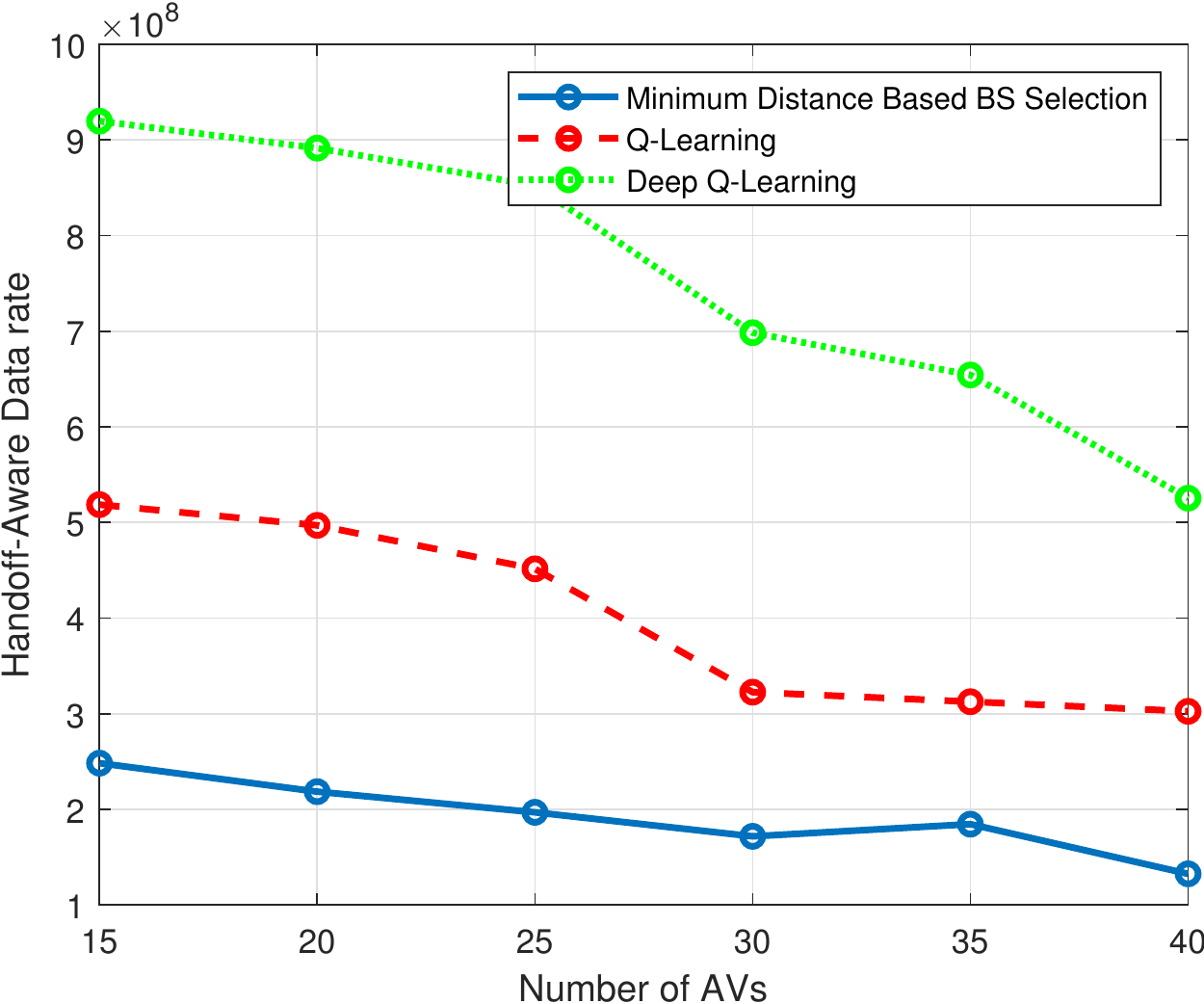}\hspace{-1cm}&\hspace{0.25cm}
\includegraphics[width=0.3\linewidth]{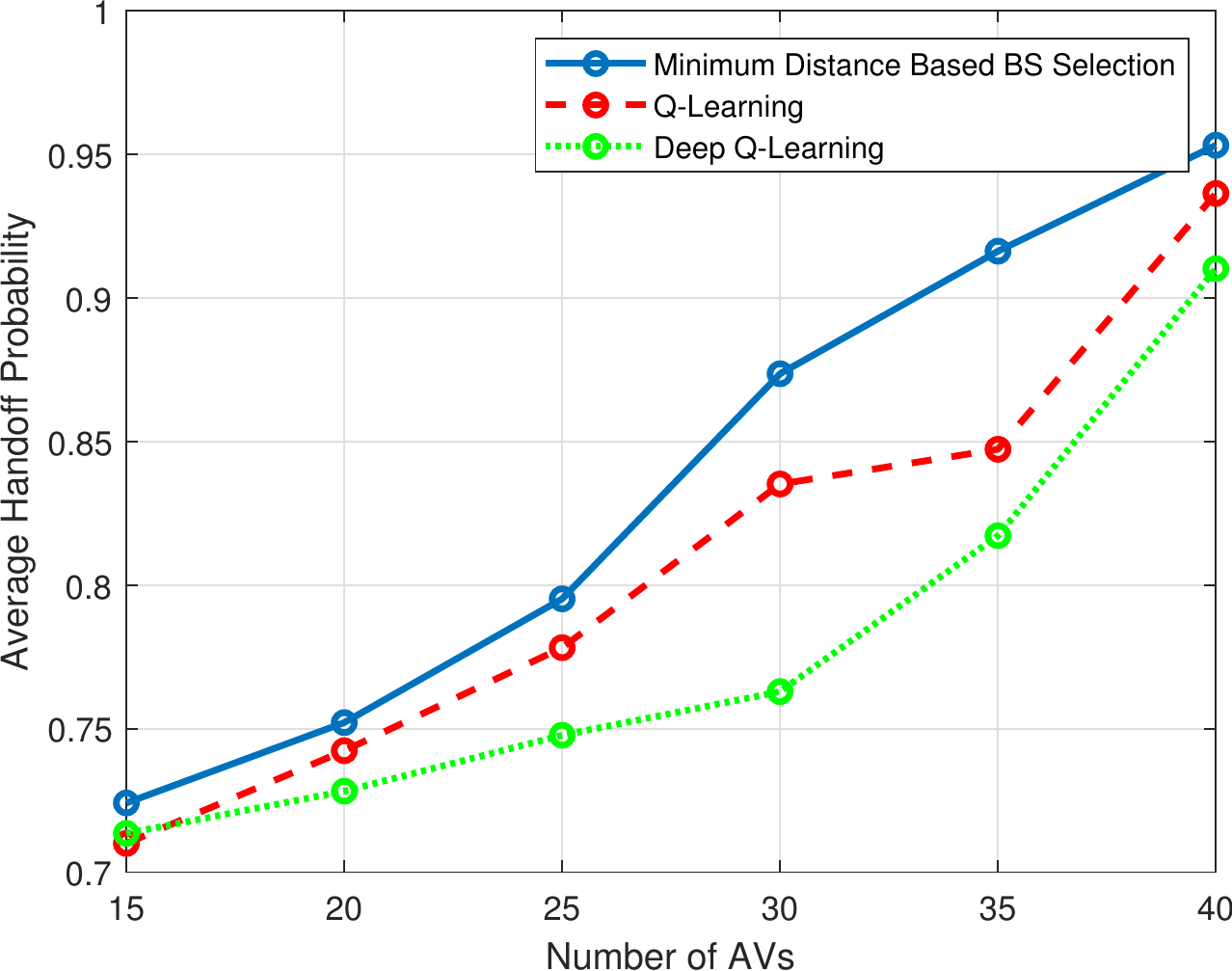}\hspace{-1cm}&\hspace{0.25cm}
\includegraphics[width=0.3\linewidth]{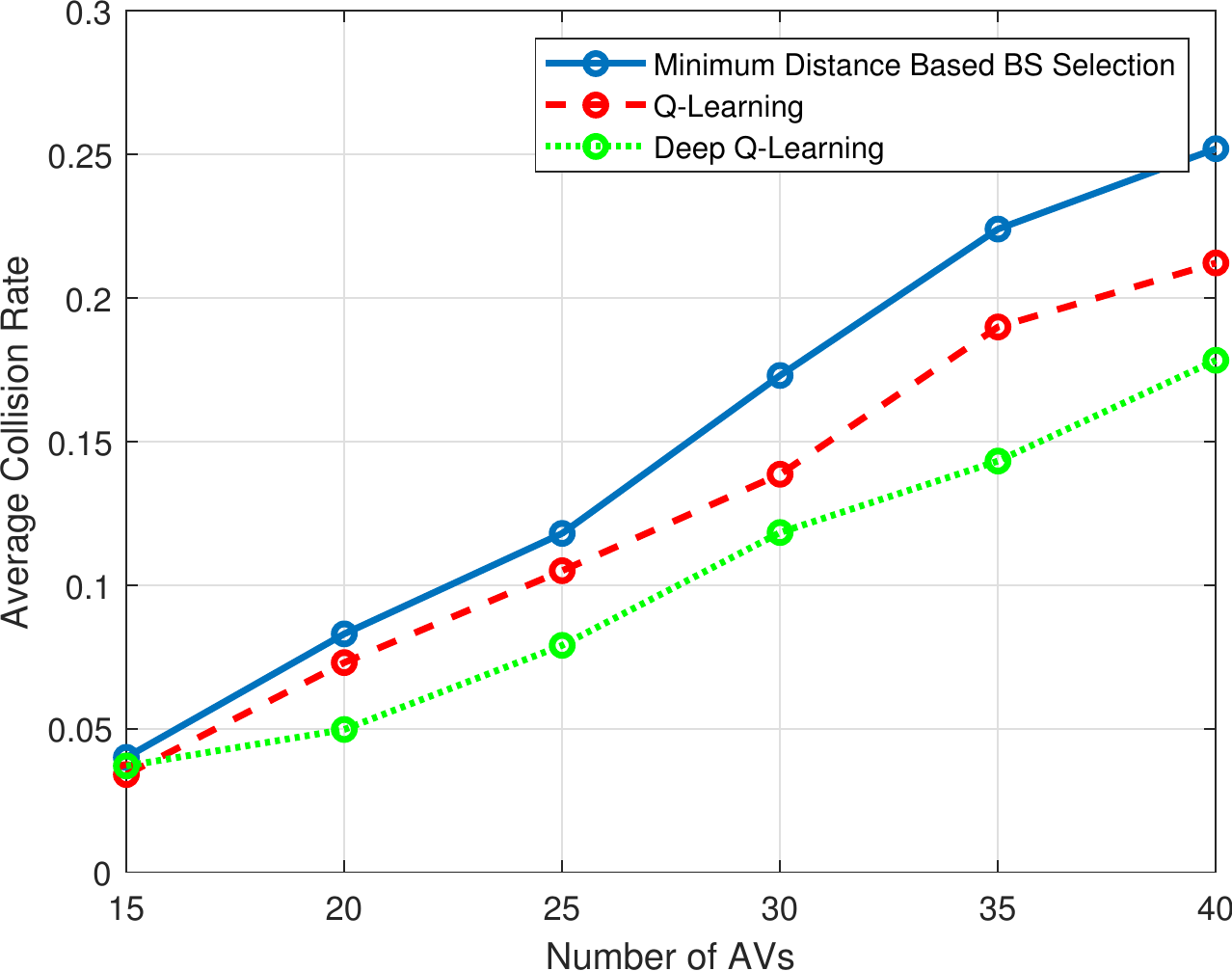}\hspace{-1cm}&\hspace{0.25cm}&\\
\qquad \qquad\qquad \qquad (a)  & (b) & (c) 
\end{tabular}
\vspace{-2mm}
\caption {(a) Handoff-aware data rate (b) Average handoff probability, (c) collision rate  considering   5 RBSs, 15 TBSs, AVs' desired velocities $v =30$m/s.} 
\label{fig4}
\end{figure*}

Fig.~\ref{fig3}  captures the handoff-aware data rate, average handoff probability, and collision rate as a function of  increasing TBSs. It also compares the performance of Q-learning, deep Q-learning, and nearest-BS selection. Evidently, deep Q-learning outperforms the other benchmarks. Also, we observe in Fig.~\ref{fig3}(a) that the handoff-aware data rate first increases due to the increase in TBSs resulting in reduced path-loss; however, after a certain point, the data rate starts degrading due to excessive handoffs experienced by the AV. The increase in handoff probability can be seen in Fig.~\ref{fig3}(b) with the increase in TBSs.  Also, we note that increasing number of BSs reduces the average achievable velocity   to minimize the handoffs and improve the handoff-aware data rate. That is, to satisfy connectivity requirements, AVs may have to sacrifice the  speeding. Subsequently, the collision rate as can be seen in Fig.~\ref{fig3}(c) reduces with the increase in BSs.


 Fig.~\ref{fig4}(a) depicts the  handoff-aware data rate, average handoff probability, and collision rate as a function of  the number of  AVs. We note that the handoff-aware data rate tends to reduce with the increase in AVs. The reason is the increase in the network traffic load at each BS.  Fig.~\ref{fig4}(b) demonstrates that handoff probability increases with the increase in AVs. The reason is that with more AVs, each AV will get minimal resources, thus AVs will likely switch to TBSs with more resources. However, TBSs are prone to handoffs. Fig.~\ref{fig4}(c) depicts that more AVs  will result in a higher  probability of a collision which is intuitive.

\bibliographystyle{IEEEtran}
\bibliography{main.bib}

\end{document}